\pdfoutput=1

\documentclass[11pt]{article}

\usepackage[final]{acl}

\usepackage{times}
\usepackage{latexsym}

\usepackage[T1]{fontenc}

\usepackage[utf8]{inputenc}

\usepackage{microtype}

\usepackage{inconsolata}

\usepackage{graphicx}

\usepackage{CJKutf8}

\newcommand{\PARASPACE}[1]{\setlength{\baselineskip}{0.7\baselineskip}#1}
\newcommand{\MYFONTSIZE}[1]{\PARASPACE{{\fontsize{8}{9}\selectfont {{#1}}}}}

\title{Can ChatGPT Really Understand Modern Chinese Poetry?}

\author{
       Shanshan Wang$^1$~~~~
        Derek F. Wong$^1\thanks{Corresponding Author}$~~~~
        Jingming Yao$^2$~~~~
        \textbf{Lidia S. Chao$^1$}~~~~
       \\
    $^1$NLP$^2$CT Lab, Department of Computer and Information Science, University of Macau \\
    {nlp2ct.shanshan@gmail.com, \{derekfw,lidiasc\}@um.edu.mo} \\
     $^2$Department of Portuguese, Faculty of Arts and Humanities, University of Macau\\
       {jmyao@um.edu.mo} \\
    }

\begin{document}
\maketitle
\begin{abstract}

ChatGPT has demonstrated remarkable capabilities on both poetry generation and translation, yet its ability to truly understand poetry remains unexplored. Previous poetry-related work merely analyzed experimental outcomes without addressing fundamental issues of comprehension. This paper introduces a comprehensive framework for evaluating ChatGPT's understanding of modern poetry. We collaborated with professional poets to evaluate ChatGPT's interpretation of modern Chinese poems by different poets along multiple dimensions. Evaluation results show that ChatGPT's interpretations align with the original poets' intents in over 73\% of the cases. However, its understanding in certain dimensions, particularly in capturing poeticity, proved to be less satisfactory. These findings highlight the effectiveness and necessity of our proposed framework. This study not only evaluates ChatGPT’s ability to understand modern poetry but also establishes a solid foundation for future research on LLMs and their application to poetry-related tasks.




\end{abstract}


\section{Introduction}



Large language models (LLMs) have been widely used for various tasks \cite{zhang2024electionsim,shen2025measuring,chatterjee2025assessing,lan-etal-2025-f2bench,ye2025unveiling,lin-etal-2025-large}. Previous studies have explored the application of ChatGPT to poetry-related tasks, primarily focusing on poetry generation \cite{antar2023effectiveness,deng2024can,hutson2023poetry}, translation \cite{virvou2023chatgpt, wang-etal-2024-best}, and detection \cite{wang-etal-2025-benchmarking}. However, without exception, these studies have drawn conclusions solely based on ChatGPT's outputs. In other words, their scope has been limited to the evaluation and analysis of experimental results. For example, \citet{virvou2023chatgpt} conducted an evaluative study examining the interpretation of poetry generated by ChatGPT. They suggest that ChatGPT has the potential to delve into the depth of poetic content. However, the fundamental question remains: Does ChatGPT truly understand poetry? Poetry, particularly modern poetry, has long faced the difficulty of being interpreted and read \cite{ZhangTaozhou2022Problems}. Despite its significance, the issue of understanding poetry has not been adequately addressed in prior research.

\begin{figure*}[t]
    \centering
    \includegraphics[width=1.0\textwidth]{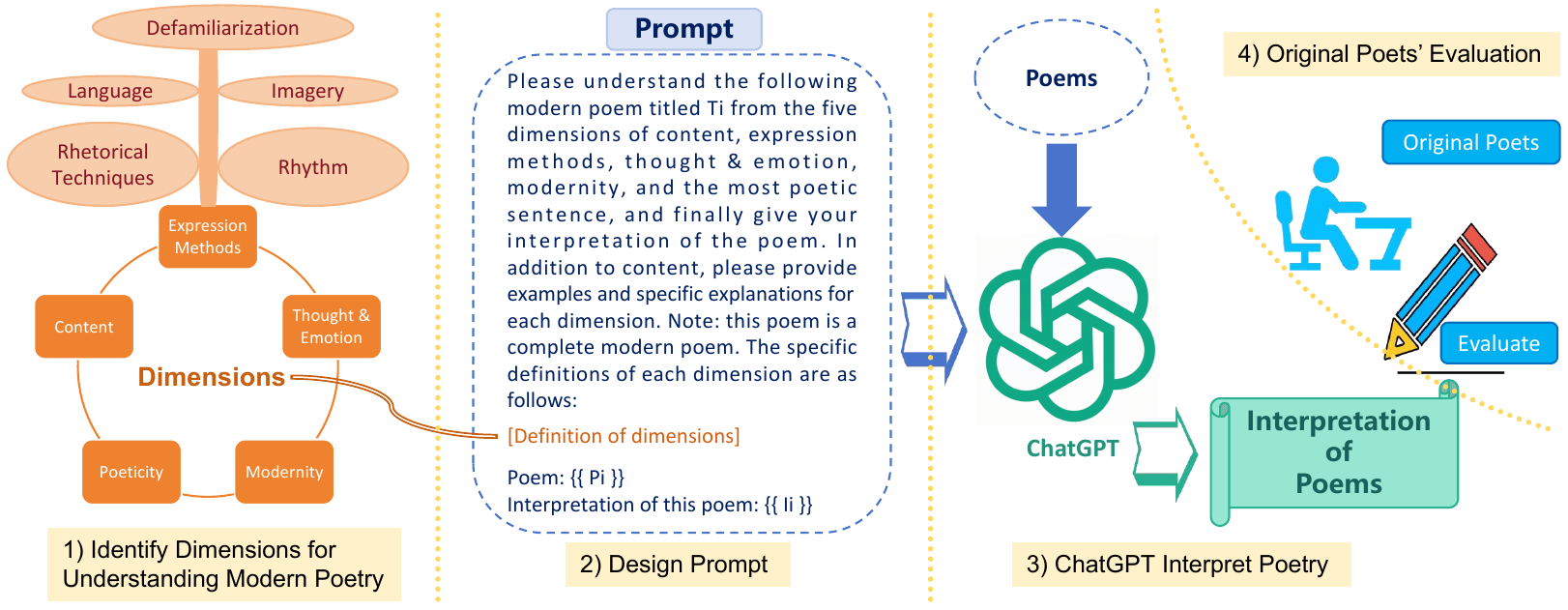}
    \caption{The framework for evaluating ChatGPT's understanding of modern poetry.}
    \label{framework}
\end{figure*}



The ability to understand poetry \cite{tate1940understanding,pierce2003defining,mcgrath2018understating} is a prerequisite for all poetry-related tasks. Without this foundation, advancements in tasks like poetry generation \cite{manurung2004evolutionary,zhang2014chinese,yi2018automatic,van2020automatic} and poetry translation \cite{genzel2010poetic,ghazvininejad2018neural,chakrabarty2021don,song-etal-2023-towards} remain constrained. Therefore, to effectively advance these tasks, it is essential to first address and emphasize the issue of ChatGPT's comprehension of poetry. In this paper, we aim to explore whether ChatGPT is capable of understanding modern poetry. Specifically, we propose a comprehensive framework for evaluating ChatGPT's understanding of modern Chinese poetry. Our approach is as follows. Firstly, in collaboration with professional poets, we identified five key dimensions crucial for understanding poetry: content, expression methods \cite{yuhaizhang2001,wangguangming1998}, thought \& emotion \cite{XiYunshuTwoCharacteristics,Emotionalclassification,WangShuting2006,WangMi2019}, modernity \cite{ChengBo2005,LongQuanming2005,ZangDi1998} and poeticity \cite{wang-etal-2024-best}. Then, we developed and refined prompts to enable ChatGPT to interpret modern poetry more accurately. Reading poetry is inherently subjective, as interpretations often vary significantly among readers \cite{Brodsky1999WitnessandPleasure, JiangYongjun2012,ZhangTaozhou2022Problems}. However, the poet's original intent provides a unique benchmark for evaluation. To assess the validity of ChatGPT's interpretations, we invited the original poets to evaluate its outputs across the identified dimensions. Our evaluation results reveal that ChatGPT's interpretations align with the poets' original intent in over 73\% of the cases. However, its performance in certain dimensions, particularly in capturing poeticity, is less satisfactory.

The primary contributions of our work are as follows: 1)  We are the first to systematically evaluate ChatGPT's ability to understand poetry. 2) We introduce a comprehensive framework to assess ChatGPT's understanding of modern Chinese poetry. 3) We define key dimensions necessary for understanding poetry and establish a systematic evaluation methodology. 4) We involved professional poets to evaluate ChatGPT's interpretations, ensuring a reliable and nuanced assessment.

\section{Evaluating ChatGPT’s Understanding of Modern Poetry}
\label{Evaluating ChatGPT’s Understanding of Modern Poetry}

To investigate whether ChatGPT can truly understand modern poetry, we propose a comprehensive framework called \textbf{ECUMP} (\textbf{E}valuation of \textbf{C}hatGPT’s \textbf{U}nderstanding of \textbf{M}odern \textbf{P}oetry). The framework, as illustrated in Figure \ref{framework}, consists of four components:




\paragraph{Identify Dimensions for Understanding Modern Poetry}



Referring to prior research on modern poetry comprehension \cite{ChangYinabrieflytalks2012} and informed by the recommendations of eight professional modern poets, we identified five critical dimensions for understanding modern poetry: 1) \textbf{Cont}ent, which refers to what the poem describes; 2) Expression Methods, which refer to the techniques employed in writing the poem. These include linguistic features (\textbf{Lang}uage), the use of visual and sensory images (\textbf{Imag}ery), rhetorical strategies such as metaphor, simile, or symbolism (\textbf{Rhet}orical Techniques), the poem's internal rhythm reflected in external forms like pauses, line breaks, and stanzas (\textbf{Rhyt}hm), and innovation through challenging conventional perceptions (\textbf{Defa}miliarization); 3) \textbf{Thou}ght \& Emotion, which captures the ideas and emotions conveyed by the poem; 4) \textbf{Mode}rnity, referring to the contemporary consciousness or modern sensibility embedded in the poem; and 5) \textbf{Poet}icity, a subjective quality encapsulating the poem's poetic essence. There may be multiple sentences in a poem with varying degrees of poetic quality, but only one sentence is considered the most poetic. To measure poeticity, we identify the most poetic sentence in the poem as a benchmark for evaluation.

\paragraph{Design Prompt}




Inspired by previous work on prompt designing \cite{jiao2023chatgpt,gao2023design} and modern poetry comprehension \cite{ZhangTaozhou2021How}, we designed and optimized a prompt to enable ChatGPT to accurately interpret modern poetry. The final prompt is presented in Table \ref{The prompt we designed}, with additional details provided in Appendix \ref{subsec: Details about Prompt Optimization}.

\begin{table}[ht]
\centering
\renewcommand{\arraystretch}{0.9}
\begin{tabular}{p{7cm}}
\hline
\MYFONTSIZE{Please understand the following modern poem titled $T_i$ from the five dimensions of content, expression methods, thought and emotion, modernity, and the most poetic sentence, and finally give your interpretation of the poem. In addition to content, please provide examples and specific explanations for each dimension. Please note that this is a complete modern poem. The specific definitions of each dimension are as follows:}  \\
\MYFONTSIZE{Content: Summarize what the poem is about.}  \\
\MYFONTSIZE{Expression Methods: Understand how this poem is written from five aspects: language, imagery, rhetorical techniques, rhythm, and defamiliarization. The five specific aspects are as follows: 1) What are the characteristics of the language of this poem? Is it innovative? 2) What imageries are used in this poem and what are their functions? 3) What rhetorical techniques is used in this poem? What is the function? 4) How is the rhythm of this poem? Such as pauses, lines, and stanzas. 5) Does this poem contain any expression of defamiliarization? If there is any unfamiliar expression, please point it out; if not, you can skip it.}  \\
\MYFONTSIZE{Thought \& Emotion: What thought and emotion does this poem express?}  \\
\MYFONTSIZE{Modernity: Is the poem modern or modern-minded?}  \\
\MYFONTSIZE{Poeticity: Choose the most poetic sentence in this poem.}  \\
\MYFONTSIZE{Poem: \{\{$P_i$\}\} (Directly used the original Chinese poem) } \\
\MYFONTSIZE{Interpretation of this poem: \{\{$I_i$\}\}}  \\

\hline

\end{tabular}
\renewcommand{\arraystretch}{1.0}
\caption{\label{The prompt we designed}The prompt we designed for ChatGPT to understand modern poetry.
}
\end{table}


\paragraph{ChatGPT Interpret Poetry}

Using the designed prompt, ChatGPT generates poems' interpretations on identified dimensions, facilitating a systematic evaluation of its comprehension ability.





\paragraph{Original Poets’ Evaluation}

Automatic evaluation methods are unsuitable for poetry-related tasks due to the unique characteristics of modern poetry \cite{refaee2023okaz,novikova2017we,wang-etal-2024-best}. Static evaluation methods may be unreliable under distribution shifts \cite{shen2026preconditioned}. Instead, we rely on human evaluation conducted by the original poets themselves. While interpretations of a poem often vary among readers \cite{JiangYongjun2012}, the author’s original intent provides a definitive benchmark. Therefore, the primary evaluation of ChatGPT’s interpretations is based on assessments by the original poets. Details of the evaluation process can be found in Section \ref{Evaluation}.



\section{Experiment}



Referring the dataset scale established in prior works \cite{wang-etal-2024-best, virvou2023chatgpt, hutson2023poetry, antar2023effectiveness}, we selected a collection of 48 modern Chinese poems for our experiments. Specifically, we collaborated with six professional modern Chinese poets, each of whom contributed eight poems they recently written. This process resulted in a high-quality dataset of 48 modern Chinese poems, comprising 147 stanzas and 750 lines in total. The dataset is divided into two categories: 40 poems from five poets that represent common types of modern poetry (denoted as Com-Poetry) and 8 poems from one poet that represent special types of modern poetry (denoted as Spe-Poetry). The Spe-Poetry category is characterized by its incorporation of content from ancient Chinese literature, making extensive use of classical imagery \cite{yeh1990new}. These poems may feature protagonists who are historical figures. Despite these characteristics, both the external forms and the underlying thoughts of Spe-Poetry align with the conventions of modern poetry, and they are therefore classified as part of this genre. Using the designed prompts, GPT-4 (gpt-4-0125) \cite{achiam2023gpt} was tasked with interpreting these 48 poems across multiple dimensions, as specified in our framework. The temperature of the model is set to \texttt{0.5}, and the top\_p is set to \texttt{1.0}. Examples of experimental output are provided in Appendix \ref{subsec: Example of the Experiment}.

\begin{table*}
  \centering
  \begin{tabular}{llllllllll}
    \hline
    &  Cont         & Lang   & Imag & Rhet
 & Rhyt & Defa  & Thou & Mode \\
    \hline
    Com-Poems & 80.33  & 79.05 & \textbf{81.18}  & 77.83  & 76.15 & 79.40 & 78.80 & 79.88 &     \\
    Spe-Poems & 77.50  & 73.75 & 81.25  & \textbf{88.75}  & 82.50 & 77.50 & 78.75 & 82.50  \\
    \hline
  \end{tabular}
  \caption{\label{40 common types of poems}
    Evaluation results of different types of poems by the original poets. 
  }
\end{table*}



\section{Evaluation}
\label{Evaluation}


\subsection{Original Poets’ Evaluation}
\label{Original Poets Evaluation}

We invited the six original poets to evaluate GPT-4’s interpretations of their own poems across five dimensions: content, expression methods, thought \& emotion, modernity, and poeticity, The definitions of these dimensions are consistent with those introduced in Section \ref{Evaluating ChatGPT’s Understanding of Modern Poetry}.






\paragraph{Evaluation Rules}

The evaluation methodology is identical for both Com-Poetry and Spe-Poetry. Each original poet evaluated only GPT-4’s interpretations of their own poems. For poeticity, scores were restricted to 0, 50, or 100 points. A score of 0 indicates that ``the most poetic sentence'' selected by GPT-4 lacks poeticity entirely, while 100 signifies that the sentence is also recognized by the original poet as the most poetic. 50 indicates that the sentence possesses poeticity but is not considered the most poetic by the original poet.




For the other four dimensions (content, expression methods, thought \& emotion, and modernity), scores ranged from 0 to 100. A score of 0 indicates that GPT-4’s interpretation of the poem is completely inconsistent with the poet’s intentions, while a score of 100 signifies complete alignment with the original poet’s thoughts. Higher scores represent a greater overlap between GPT-4’s understanding and the original poet’s intent.

\paragraph{Evaluation Results}

\begin{figure}[t]
\centering
    \includegraphics[width=\columnwidth]{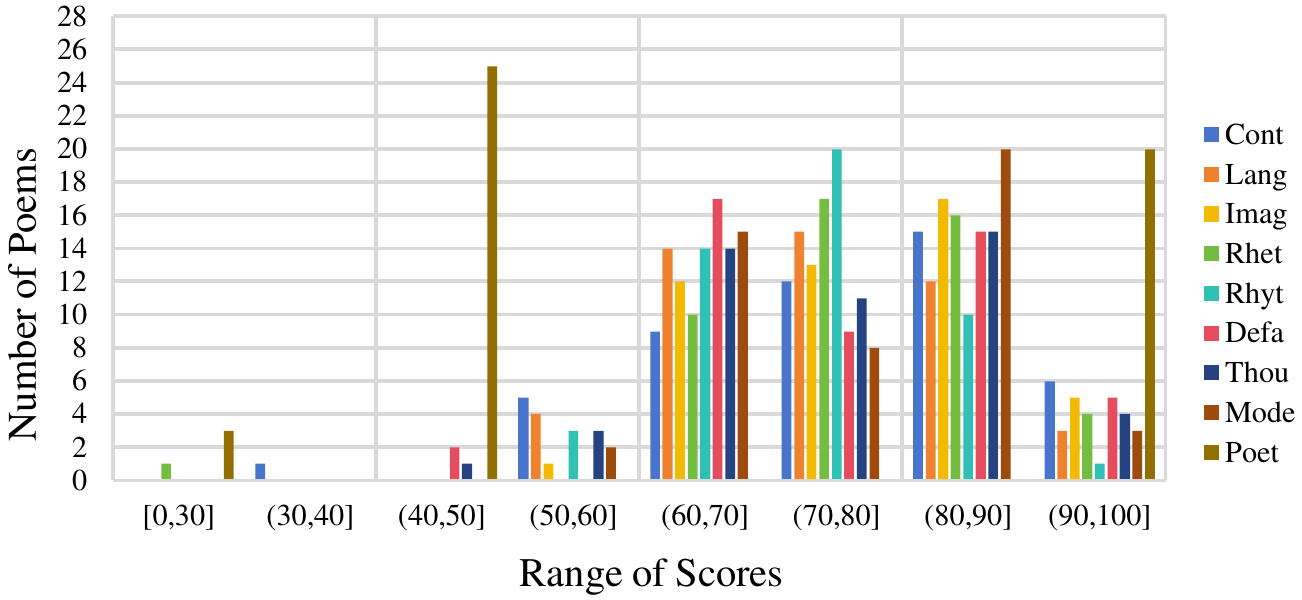}
    \caption{The distribution of evaluation scores for GPT-4’s interpretation of 48 poems.}
    \label{figure of 40 poems}
\end{figure}

Table \ref{40 common types of poems} presents the evaluation results provided by the six original poets on GPT-4’s interpretation of 48 poems across all dimensions except poeticity. Table \ref{Evaluation results of poeticity} summarizes the poets’ evaluations of GPT-4’s understanding of poeticity. Figure \ref{figure of 40 poems} illustrates the distribution of scores for GPT-4’s interpretation of 48 poems, evaluated by six poets across different dimensions, with a focus on the range of scores.






\begin{table}
  \centering
  \begin{tabular}{llllllllll}
    \hline
  & N-100  & N-0   & N-50 \\
    \hline
    Com-Poems   & 17   & 2  & 21   \\
    Spe-Poems   & 3  & 1 & 4  \\
    \hline
  \end{tabular}
  \caption{\label{Evaluation results of poeticity}
    Evaluation results of ChatGPT's understanding of poeticity. ``N-100'', ``N-0'' and ``N-50'' respectively represent the number of occurrences of 100, 0, and 50.
  }
\end{table}



\subsection{LLMs’ Evaluation}



Traditional automatic evaluation methods are not well-suited for poetry-related tasks due to the unique characteristics of poetry \cite{novikova2017we, refaee2023okaz, wang-etal-2024-best}. Fortunately, LLMs were used as evaluation tools due to their obvious potential and advantages \cite{chiang2023can, wang2023chatgpt, liu2023g}, and have made effective progress in the evaluation of multiple tasks such as open-domain conversations \cite{lin2023llm}, text summarization \cite{chu2024pre}, dialogue response generation, and open-ended question answer \cite{chan2023chateval}. Therefore, we use LLMs as evaluation tools to evaluate the interpretations generated by GPT-4, applying the same evaluation rules as human evaluators. Table \ref{The Prompt we designed for the LLMs (without enhancement) as the evaluators} presents the prompt we designed for LLMs as the evaluators to evaluate GPT-4’s understanding of poems.
Table \ref{Evaluation results for GPT-4's understanding of Com-Poems by LLMs} and Table \ref{Evaluation results for GPT-4's understanding of Spe-Poems by LLMs} present the evaluation results of LLMs as the evaluators on GPT-4's understanding of Com-Poems and Spe-Poems, respectively.

\begin{table}[t]
\small
\begin{tabular}{p{7.0cm}} 
\hline
Please score the following ``readers' understanding of the poem'' from the eight dimensions of content, language, imagery, rhetorical techniques, rhythm, defamiliarization, thought and emotion, and modernity, and give detailed reasons for your score. Please note that you only need to score and give reasons for the ``Readers' Understanding of this poem, not the poem itself.  \\
\vspace{0.001cm}
\textbf{Scoring Dimensions:} Same as human evaluation. \\
\vspace{0.001cm}
\textbf{Scoring Rules:} Each dimension ranges from 0 to 100. A score of 0 means that you think the reader's understanding of a certain dimension is completely wrong, and 100 indicates that you think the reader's understanding of a certain dimension is completely correct; the more correct the reader's understanding of a certain dimension is, the higher the score you will give.  \\
\vspace{0.001cm}
\textbf{Poem:} \{\{$P_i$\}\}   \\
\vspace{0.001cm}
\textbf{Readers' Understanding of this Poem:} \{\{$U_i$\}\}  \\
\hline
\end{tabular}
\caption{\label{The Prompt we designed for the LLMs (without enhancement) as the evaluators}The prompt we designed for the LLMs as the evaluators to evaluate GPT-4’s understanding of poems.}
\end{table}



\begin{table*}
  \centering
  \begin{tabular}{lllllllll}
    \hline
   & Cont         & Lang   & Imag & Rhet
 & Rhyt & Defa  & Thou & Mode \\
    \hline
   
   o1-preview &  94.70	& 92.03	& 96.33	& 91.40	& 90.35	& 93.90	& 95.23	& 96.75 \\
       
   GPT-4o & 92.38	& 89.38	& 94.00	& 88.50	& 88.00	& 87.38	& 91.75	& 93.25 \\
    

   
   Qwen2.5-Plus &  94.50	& 89.50	& 94.50	& 90.00	& 85.13	& 91.25	& 90.25	& 94.88  \\
   
   Qwen2.5-Max & 95.00	& 89.13	& 94.13	& 93.00	& 90.00	& 86.13	& 95.00	& 95.00  \\


    \hline
  \end{tabular}
  \caption{\label{Evaluation results for GPT-4's understanding of Com-Poems by LLMs}
    The evaluation results of LLMs as the evaluators on GPT-4's understanding of Com-Poems.
  }
\end{table*}

\begin{table*}
  \centering
  \begin{tabular}{llllllllll}
    \hline
   &  Cont         & Lang   & Imag & Rhet
 & Rhyt & Defa & Thou & Mode \\
    \hline
    

    o1-preview & 95.00	& 92.50	& 96.25	& 88.75	& 93.13	& 83.13	& 96.25	& 94.38 \\
    
    GPT-4o & 91.88	& 91.25	& 94.38	& 87.50	& 93.75	& 81.88	& 92.50	& 93.75 \\

    Qwen2.5-Plus & 95.00	& 90.00	& 95.00	& 90.00	& 85.00	& 91.25	& 90.00	& 95.00 \\

    Qwen2.5-Max & 95.00	& 90.00	& 95.00	& 90.00	& 90.00	& 85.00	& 95.00	& 95.00 \\

    \hline
  \end{tabular}
  \caption{\label{Evaluation results for GPT-4's understanding of Spe-Poems by LLMs}
    The evaluation results of LLMs as the evaluators on GPT-4's understanding of Spe-Poems.
  }
\end{table*}

\section{Analysis}

\paragraph{Human Evaluation}
Table \ref{40 common types of poems} and Figure \ref{figure of 40 poems} show that, except for poeticity, GPT-4 demonstrates a strong understanding of poetry across all dimensions, achieving at least 73\% alignment with the original poets' intentions. And the scores are concentrated between 60 and 90, indicating a reasonably consistent performance. However, GPT-4’s understanding of poeticity is unsatisfactory. When tasked with identifying the most poetic sentences, GPT-4’s responses (28 sentences) were incorrect in more than half of the cases (Table \ref{Evaluation results of poeticity}). Notably, GPT-4 selected three sentences that were entirely devoid of poeticity, highlighting significant limitations in its ability to discern poetic essence. For Com-Poems, GPT-4 demonstrates the strongest understanding of imagery, achieving an average score of 81.18 (Table \ref{40 common types of poems}). This result aligns with expectations, as imagery is a salient and explicit feature of poetry. Additionally, GPT-4 is capable of accurately grasping much of the poem’s content, with a score of 80.33. Notably, GPT-4 performs better with poems that have clear themes or narrative structures, providing more comprehensive and accurate summaries with fewer omissions. For Spe-Poems, GPT-4’s performance diverges significantly from its results with Com-Poems. It demonstrates the strongest understanding of rhetorical techniques (88.75), rhythm (82.50), and modernity (82.50). For example, GPT-4 is generally able to identify pauses in Spe-Poems and analyze them in ways that align closely with the original poets’ intentions. However, GPT-4 struggles with language in Spe-Poems, scoring the lowest (73.75) in this dimension. While it can recognize that some of the language originates from ancient Chinese literature, the inherent complexity of Spe-Poems poses challenges to GPT-4’s ability to fully comprehend the nuances of the language. This aligns with previous works \cite{lan2025mcbe, LyuBias, QiBias} that the cultural context behind a text influences the performance of LLMs. 

There are several key insights emerge from the analysis. First, in addition to identifying external imagery, GPT-4 can detect the implicit modern consciousness embedded within the poems. Second, GPT-4 performs better when the language expression is straightforward, with greater coherence and stronger correlations between words leading to more accurate comprehension. Third, GPT-4 can reliably identify similes in sentences with explicit cues such as ``like,'' ``as,'' and ``is.'' However, its ability to interpret rhetorical techniques diminishes significantly, or even fails entirely, when such explicit cues are absent. More detailed analyses are provided in Appendix \ref{subsec: Detailed Analysis}.

\paragraph{LLMs Evaluation}
Disappointingly, the results from LLMs deviated significantly from the evaluations provided by professional poets, indicating that LLMs are not yet reliable for evaluating the understanding of poetry. For Com-Poems, although GPT-4o performed slightly closer to human evaluations compared to other models, all LLMs' results were still significantly different from those of the professional poets. This finding highlights the necessity and reliability of human evaluation in tasks requiring nuanced and culturally embedded understanding, such as poetry.

Our findings further highlight the reliability and necessity of human evaluation in poetry domain. Poetry is inherently nuanced, with elements such as imagery, rhythm, and thought requiring a deep cultural and contextual understanding that current AI models cannot fully replicate. As such, human evaluation remains indispensable for accurately assessing poetic comprehension in our study.

\section{Conclusion}

Understanding poetry is a fundamental prerequisite for all poetry-related tasks. In this paper, we propose a comprehensive framework to evaluate ChatGPT’s ability to understand modern poetry and examine the extent of its comprehension. Specifically, we identify several critical dimensions essential for understanding poetry, informed by professional expertise in the field. Using these dimensions, ChatGPT was tasked with interpreting the poems provided by various poets, based on carefully designed prompts. To assess these interpretations’ quality, we invited the original poets to evaluate GPT’s outputs across the identified dimensions. The evaluation results show that over 73\% of ChatGPT’s interpretations align with the original poets’ intentions and thoughts. However, significant room for improvement remains in specific dimensions, particularly poeticity. These findings highlight the effectiveness and necessity of our proposed framework. This study establishes a solid foundation for future research on LLMs and their application to poetry-related tasks.

\section*{Limitations}
The data we utilized consists of recently written poems by professional poets, so there has been no prior human analysis of these poems to date. Consequently, in our study, the evaluation of ChatGPT's poetic comprehension was conducted manually by the original poets themselves, which required a high cost. Fortunately, our work substantiates the effectiveness of ChatGPT as a tool to assist poetry comprehension, which lays a solid foundation for future research on LLMs related to poetry.




\section*{Ethical Considerations}
Our work has validated the effectiveness of ChatGPT as a tool for assisting in the understanding of poetry, as it is capable of providing human readers with interpretations that largely align with the original poets' intentions. However, this could potentially lead to an over-reliance on ChatGPT's interpretations by human readers.

\section*{Acknowledgements}


This work was supported in part by the Science and Technology Development Fund of Macau SAR (Grant No. FDCT/0007/2024/AKP), the UM and UMDF (Grant Nos. MYRG-GRG2024-00165-FST-UMDF, MYRG-GRG2025-00236-FST, EF2024-00185-FST), the Tencent AI Lab (EF2023-00151-FST), the Stanley Ho Medical Development Foundation (Grant No. SHMDF-AI/2026/001), and the National Natural Science Foundation of China (Grant No. 62266013).

\bibliography{main}

\appendix

\section{Appendix}
\label{sec: appendix}

\subsection{Professional Poets}
\label{subsec: Professional Poets}
 
The poets involved in this work are not only poetry creators but also seasoned researchers of poetry theory, including members of the Chinese Writers Association\footnote{https://www.chinawriter.com.cn/n1/2025/0417/c403937-40462301.html}, poetry editors, and university professors. They are all excellent Chinese poets who have published many poetry collections, published many articles on modern Chinese poetry, and won many poetry awards. 




\subsection{Details About Prompt Optimization} 
\label{subsec: Details about Prompt Optimization}
During the experimental process, we observed that when GPT-4 generated the interpretation of poetry, a reminder ``You did not provide the complete text of the poem'' appeared before the interpretation. To quantify this occurrence, we tallied the frequency of such prompts and discovered that in the analysis of 48 poems, more than one-third of the cases (18 poems in total) elicited a similar response. This suggests that GPT-4 may not fully understand the completeness of poetry. To better stimulate GPT-4’s understanding of poetry, we added the content ``This is a complete modern poem'' to the designed prompt. Finally, all experiments used the optimized prompts 
shown in Table \ref{The prompt we designed}.

\subsection{Example of the Experiment}
\label{subsec: Example of the Experiment}

Here is an example of the experiment on GPT-4.


\textbf{Poem:}

\begin{center}
\begin{CJK}{UTF8}{gkai}醉酒一事\end{CJK}

(The Matter of Being Drunk)\\\

\begin{CJK}{UTF8}{gkai}在酒的国度里，鱼鳌可以\end{CJK}

(In the kingdom of wine, turtles might)

\begin{CJK}{UTF8}{gkai}离开海洋，鸟雀不必站在树上\end{CJK}

(Depart the sea, and birds need not perch in trees)

\begin{CJK}{UTF8}{gkai}男人醉酒后，很放心地\end{CJK}

(A man, inebriated, with carefree trust)

\begin{CJK}{UTF8}{gkai}把自己摆到了街边。人间之事\end{CJK}

(Lay himself along the street’s edge, where worldly matters)

\begin{CJK}{UTF8}{gkai}全都碎在杯里，一饮而尽后\end{CJK}

(All shattered in the cup, drained in one gulp)

\begin{CJK}{UTF8}{gkai}在肠胃里激荡，像巨浪冲刷着堤防\end{CJK}

(Surging in the stomach, like giant waves washing over the embankment)\\\

\begin{CJK}{UTF8}{gkai}车辆在街灯的重影里轰鸣、拉扯\end{CJK}

(Vehicles roar and tug in the doubled shadows of streetlights)

\begin{CJK}{UTF8}{gkai}犹如那些轻易给出的诺言，在黑暗中\end{CJK}

(Like promises too lightly made, in dark)

\begin{CJK}{UTF8}{gkai}徒劳地撒网，没人会在意一个醉汉\end{CJK}

(Casting nets in vain. No one cares about a drunk)

\begin{CJK}{UTF8}{gkai}倒置的身体和语言，他只是马路上的一块礁石\end{CJK}

(Inverted body and language, he is just a rock on the road)

\begin{CJK}{UTF8}{gkai}是礁石上一条搁浅的鱼，鱼腹里的一根刺\end{CJK}

(It's a stranded fish on the reef, a thorn in its belly)\\\

\begin{CJK}{UTF8}{gkai}想到这里，他真的像一根鱼刺一样\end{CJK}

(Thinking of this, he really looks like a fishbone)

\begin{CJK}{UTF8}{gkai}卡住了道路的喉咙\end{CJK}

(Stuck in the throat of the road)
\end{center}

\textbf{Interpretation of this poem by GPT-4:}

\begin{itemize}

\item \textbf{Content:}  
This poem describes the state and feelings of a drunkard. It depicts how alcohol allows one to escape reality, making the drunkard feel as if they are in a ``kingdom of wine'' where they are not bound by the rules of the real world. However, this escapism ultimately renders him a mere rock in the city, a stranded fish, even a fishbone stuck in the throat of the road.

\item \textbf{Expression Methods:}
\textit{1) Language Characteristics.} The poem uses direct and concrete language, innovatively comparing the drunkard to a rock and a fishbone, which is a common language feature in modern poetry.
\textit{2) Imagery.} The poem uses images like ``turtles,'' ``birds,'' ``giant waves,'' and ``vehicles,'' comparing the state of drunkenness to the freedom of animals and the power of natural disasters, which enhances the expressiveness of the poem.
\textit{3) Rhetorical Techniques.} The poem employs rhetorical devices such as metaphor (comparing the drunkard to a rock and fishbone) and hyperbole (e.g., ``all shattered in the cup''), which enhance the expressiveness and emotional impact of the language.
\textit{4) Rhythm.} The poem controls the rhythm through short lines and pauses, simulating the discontinuous stream of consciousness of the drunkard and his unsteady presence in the city.
\textit{5) Defamiliarization.} The poem achieves a defamiliarizing effect by comparing people to objects (rock, fishbone), allowing readers to see the image and state of the drunkard from a new perspective.

\item \textbf{Thought \& Emotion:}
The poem conveys a deep understanding of the state of drunkenness and sympathy for the isolated and unsupported state of the drunkard in society. Through the image of the drunkard, the poet reflects the potential for people to become lost and lonely in modern society.

\item \textbf{Modernity:}
This poem has modernity; it takes the experience of an ordinary person in modern society as a starting point, using modern poetic language and expression to explore the individual's isolated state in society and the complexity of the inner world. Through the portrayal of the drunkard, the poet reflects the inner struggle of modern people and their longing for freedom.

\item \textbf{Poeticity:}
The sentence ``Surging in the stomach, like giant waves washing over the embankment'' is particularly poetic. It vividly describes the effect of alcohol in the body and metaphorically alludes to the inner turmoil and conflict within a person.

\end{itemize}

\subsection{Detailed Analysis}
\label{subsec: Detailed Analysis}


\paragraph{Com-Poems} Detailed analysis of Com-Poems across dimensions (From highest to lowest performance): 

\textbf{1) Imagery}. GPT-4 achieved the highest comprehension score in the imagery dimension, with an average score of 81.18 (Table \ref{40 common types of poems}). This aligns with expectations, as imagery is a salient and explicit characteristic of poetry. Notably, GPT-4 scored above 80 in imagery interpretation 18 times (Figure \ref{figure of 40 poems}), demonstrating its relatively strong capability in recognizing and interpreting visual and descriptive elements in poetry. This finding highlights GPT-4's ability to engage with one of the most concrete and explicit dimensions of poetic language.

\textbf{2) Content}. GPT-4 scored 80.33 in content comprehension, indicating its capability to discern the core subject matter and themes of poems. We observed that GPT-4 performs particularly well with poems that have clear themes or straightforward narrative structures, often generating comprehensive summaries with minimal omissions. For instance, GPT-4 scored above 80 in content understanding 19 times (14 scores between 80–90 and 5 scores above 90), the highest frequency among dimensions with scores above 80 (Figure \ref{figure of 40 poems}). This finding underscores GPT-4's proficiency in grasping the explicit content of modern poetry.

\textbf{3) Modernity}. Scoring 79.88, GPT-4 demonstrated a strong ability to recognize modernity, particularly through the imagery unique to modern Chinese poetry. While external modern imagery is relatively explicit, we were surprised to find that GPT-4 also identified implicit modern consciousness embedded within the poems. This nuanced capability reflects its potential to go beyond surface-level comprehension when analyzing modern poetic elements.

\textbf{4) Defamiliarization}. GPT-4 scored 79.40 in defamiliarization, placing it between modernity and language. Defamiliarization, often expressed through innovative linguistic techniques and modernist stylistics, was recognized by GPT-4 in a manner consistent with established poetic analysis. This suggests that GPT-4 can reasonably interpret creative deviations in language and style, even if its understanding of such techniques is not perfect.

\textbf{5) Language}. GPT-4 scored 79.05 in language comprehension. While this score is relatively high, our analysis revealed limitations in its interpretative depth. GPT-4 often resorts to generalized phrases when analyzing linguistic features, which, while broadly applicable, lack specificity for individual poems. For example, GPT-4 frequently describes the language characteristics of modern poetry as employing ``succinct and direct modern Chinese language'' with ``a straightforward and unembellished lexicon.'' These descriptions are accurate at a general level but fail to capture the unique linguistic features of specific poems. Additionally, we found that GPT-4's comprehension improves with simpler and more coherent language. When linguistic leaps are minimal and word correlations are stronger, GPT-4's interpretations are more precise.

\textbf{6) Thought \& Emotion}. GPT-4 scored 78.80 in thought \& emotion. We find that GPT-4 is adept at capturing emotional undertones in poems with clear and vivid imagery. However, its performance declines when themes become more complex or abstract, indicating a limitation in its ability to grasp deeper cognitive and emotional nuances. This finding emphasizes the critical role of imagery in facilitating GPT-4's understanding of a poem's emotional and intellectual depth.


\textbf{7) Rhetorical Techniques}. GPT-4 scored 77.83 in rhetorical techniques, slightly outperforming its understanding of rhythm. Our analysis shows that GPT-4 reliably identifies similes with explicit cues (e.g., ``like,'' ``as,'' ``is''). However, its ability to recognize more subtle or implicit rhetorical devices is significantly weaker, often leading to incorrect interpretations. For example, GPT-4 misidentified the line ``When your message arrives on my phone, I hear your voice'' as employing personification. This interpretation is incorrect because: the line does not attribute human-like qualities to inanimate objects; the relationship between ``message'' and ``voice'' is sequential and causal, not rhetorical.



Such errors highlight GPT-4's limitations in detecting rhetorical techniques without explicit cues.

\textbf{8) Rhythm}. GPT-4's comprehension of rhythm scored a relatively low 76.15. We observed that GPT-4 lacks the ability to perform a detailed rhythmic analysis. Instead, it defaults to generalized observations, such as noting variations in sentence length, pauses, and line breaks as indicators of rhythm. Notably, GPT-4 scored above 80 in rhythm comprehension only 8 times, while scores at or below 70 accounted for 16 occurrences. This finding indicates that GPT-4 struggles to capture the nuanced rhythmic patterns that are integral to poetic expression.

\textbf{9) Poeticity}. Poeticity was the weakest dimension, with GPT-4 scoring the lowest overall (Table \ref{Evaluation results of poeticity}). In the task of identifying the most poetic sentence from 40 poems, GPT-4's selections were incorrect for more than half the cases (23 sentences). For example, GPT-4 misidentified the phrase ``tempering a rod of iron into a needle'' as highly poetic, interpreting it as a metaphor for dedication and refinement. While this interpretation is valid in isolation, GPT-4 failed to consider the line's context within the poem, leading to an erroneous judgment of poeticity. Each poem is a vibrant entity \cite{ZhangTaozhou2021How}, where lines within the poem are both independent and yet intimately interconnected. This suggests that GPT-4 tends to analyze sentences in isolation rather than considering their interplay within the holistic structure.


\paragraph{Spe-Poems}

Within our dataset, Spe-Poetry represents a unique type of modern poetry, often incorporating references to ancient literary content. Prior to conducting the experiments, we anticipated that GPT-4 might struggle with understanding this type of poetry, but the experimental results were surprising, revealing both strengths and limitations. For Spe-Poems, detailed findings for key dimensions are as follows:

\textbf{1) Rhetorical Techniques}. For Spe-Poems, GPT-4 demonstrated the strongest understanding of rhetorical techniques, with an impressive agreement rate of 88.75\% with the original poets' ideas, the highest score across all dimensions. This result is particularly noteworthy. For example, GPT-4 correctly identified the use of metaphor in the line, ``The undulating mountain ranges resemble a giant hastening along his way,'' accurately recognizing the explicit cue word ``resemble'' and providing an error-free analysis of the rhetorical device. This finding is consistent with GPT-4's performance in common modern poetry, where it excels in recognizing rhetorical techniques with explicit signals. The high accuracy in this dimension highlights GPT-4's capability to analyze and interpret rhetorical elements in poetry where cues are explicit.

\textbf{2) Modernity}. GPT-4's comprehension of modernity in Spe-Poems reached 82.50\% agreement with the original poets' intentions, which was a surprising and encouraging result. For instance, in the poem ``Lotus Leaves as Clothing (I),'' GPT-4 correctly recognized the modern elements interwoven with ancient themes. While the poem is centered on the ancient poet Qu Yuan, it incorporates modern elements such as the mention of ``296 BC'' and juxtaposes ancient imagery with modern psychological experiences. GPT-4 successfully identified the poem's reflection of universal human emotions, such as loneliness, confusion, and the pursuit of beauty, that transcend time and space, effectively interpreting the modern consciousness embedded within the poem. This demonstrates GPT-4's potential to bridge classical and modern elements in poetic interpretation.

\textbf{3) Rhythm}. GPT-4's understanding of rhythm in Spe-Poems scored 82.50, second only to rhetorical techniques. This result stands in contrast to its relatively weaker performance in rhythm comprehension for common types of modern poetry. For example, GPT-4 accurately analyzed the rhythm of ``Lotus Leaves as Clothing (I),'' recognizing how line breaks and pauses, such as in the brief sentences ``Night. The lodge.'', create a tranquil and solitary atmosphere. This analysis was entirely consistent with the original poet's intent, demonstrating that GPT-4 can effectively recognize rhythmic features when they are explicitly designed to evoke particular moods or tones.

\textbf{4) Language}. GPT-4’s alignment with the original poets in language comprehension for Spe-Poems was 73.75\%, the lowest among all dimensions. This lower score can be attributed to the inherent challenges posed by this category of poetry, which often employs classical Chinese imagery and lexicon. These elements are laden with rich cultural connotations and intrinsic meanings that require a deep understanding of the relevant cultural and historical context. For example, in analyzing the poem ``Lotus Leaves as Clothing (I),'' GPT-4 correctly observed the use of a classical poetic style, accurately identifying ancient phrases like ``huai xi'' while also recognizing the integration of modern language. However, GPT-4 mistakenly categorized ``wrinkles and late spring'' as modern language, despite their frequent use as poetic images in ancient Chinese poetry. While GPT-4 correctly identified the innovation in the poem, it misunderstood that the sentence ``Qu Yuan carrying wrinkles and late spring'' innovatively materializes ``wrinkles and late spring'' through the word "carrying,'' blending modern techniques with classical imagery. Although GPT-4 struggles with fully grasping the linguistic nuances of classical elements, it exhibits the potential to identify their presence and stylistic relevance. This suggests that its understanding of language in Spe-Poems, while imperfect, demonstrates a promising foundation for further improvement.



\textbf{5) Poeticity}. GPT-4's ability to understand poeticity in Spe-Poems remains unsatisfactory, consistent with its performance on common modern poetry. When tasked with identifying the most poetic sentence from 8 poems, GPT-4's selections were incorrect in more than half the cases (5 out of 8). In one instance, it selected a sentence entirely devoid of poetic sense as the most poetic. This finding reinforces GPT-4's tendency to analyze sentences in isolation, overlooking their contextual interplay within the broader structure of a poem. This limitation highlights an area where future improvements are necessary to enhance GPT-4's capacity for holistic poetic interpretation.

The detailed analysis presented above demonstrates the analytical depth of our work. By systematically evaluating GPT-4's performance across multiple poetic dimensions and providing specific examples, we offer nuanced insights into its strengths and limitations when analyzing Com-Poems and Spe-Poems.

\end{document}